\documentclass[conference]{IEEEtran}
\IEEEoverridecommandlockouts
\usepackage{cite}
\usepackage{amsmath,amssymb,amsfonts}
\usepackage{algorithmic}
\usepackage{graphicx}
\usepackage{textcomp}
\usepackage{xcolor}
\def\BibTeX{{\rm B\kern-.05em{\sc i\kern-.025em b}\kern-.08em
    T\kern-.1667em\lower.7ex\hbox{E}\kern-.125emX}}
\begin{document}

\title{Is your forecaster smarter than an energy engineer: a deep dive into electricity price forecasting}

\author{\IEEEauthorblockN{Maria Margarida Mascarenhas}
\IEEEauthorblockA{\textit{ELECTA, KU Leuven} \\
Belgium \\
mariamargaridademendocadeataydep.d@student.kuleuven.be}
\and
\IEEEauthorblockN{Hussain Kazmi}
\IEEEauthorblockA{\textit{ELECTA, KU Leuven} \\
Belgium\\
hussainsyed.kazmi@kuleuven.be}
}

\maketitle

\begin{abstract}

The field of electricity price forecasting has seen significant advances in the last years, including the development of new, more accurate forecast models. These models leverage statistical relationships in previously observed data to predict the future; however, there is a lack of analysis explaining these models, which limits their real world applicability in critical infrastructure. In this paper, using data from the Belgian electricity markets, we explore a state-of-the-art forecasting model to understand if its predictions can be trusted in more general settings than the limited context it is trained in. If the model produces poor predictions in extreme conditions or if its predictions are inconsistent with reality, it cannot be relied upon in real-world where these forecasts are used in downstream decision-making activities. Our results show that, despite being largely accurate enough in general, even state of the art forecasts struggle with remaining consistent with reality.

\end{abstract}

\begin{IEEEkeywords}
electricity price forecasting, LEAR, explainability models
\end{IEEEkeywords}

\section{Introduction}

With the liberalization of the European electricity system, market prices serve several important purposes. On the one hand, they allow grid operators to promote competition and ensure grid stability. On the other, they allow market participants, such as generators and demand aggregators etc., to maximize their profits through new services and algorithmic innovations. In other words, they allow scheduling algorithms to determine the optimal time to consume or produce electricity \cite{b1}.

Forecasting electricity prices is therefore becoming increasingly important due to the competition in the market. An accurate forecasting algorithm therefore increases the utility of relevant stakeholders.
In this paper, we investigate the behavior of a state-of-the-art electricity price forecasting algorithm \cite{b2}. More concretely, we address the question of whether this forecaster is physically consistent, i.e. whether it attributes the right importance to the relevant input features when making forecasts for the future, and if its predictions can be trusted in extreme market conditions? For dependable and mission critical infrastructure, it may actually be preferable to have a slightly less accurate model that makes its predictions giving a physically consistent importance to input features, than a more accurate model that makes more accurate predictions for dubious reasons. The latter model will often fail to generalize in unexpected conditions, such as the ones that arose in the aftermath of pandemic-related disruptions.

\section{Data and method}

We investigate the performance of a state of the art price forecaster (LEAR model) for the EPEX-BE electricity market. The EPEX-BE market prices demonstrate high volatility with several spikes and negative prices. Section \ref{AAA} shows a detailed list with the features used for this market to make the price forecasts, and where they were obtained. Section \ref{AAC} presents a detailed description of the forecast model.

\subsection{Data}\label{AAA}

The data used by the forecaster to predict  EPEX-BE spot prices was obtained from the ENTSO-E transparency platform website \cite{b5} for the Belgium features, and the Rte website \cite{b6} for the French features. The data presented below are the input features used by the forecast model ranging from 01/01/2015 until 31/12/2021.

\begin{itemize}
    \item Electricity day-ahead prices
    \item Day-ahead generation forecast (FR)
    \item Day-ahead system load forecast (FR)
    \item Day-Ahead generation forecast for wind (BE)
    \item Day-Ahead generation forecast for solar (BE)
    \item Day-ahead system load forecast (BE)
\end{itemize}

There were some missing days in these fields, which were substituted with the average of the day before and after the missing data.

\subsection{The LEAR Model}\label{AAC}

The LEAR model uses past values as predictors and has exogenous variables as explanatory variables.
LEAR uses the least absolute shrinkage and selection operator (LASSO), a feature selection algorithm, to choose among the variables and estimate their parameters \cite{b7}. The model has previously been shown to offer excellent accuracy to performance ratio \cite{b2}. The input features that the model needs to predict the 24 day-ahead prices for day \(d\) (\(p_d=[p_{d,1},...,p_{d,24}]\)) include:

\begin{itemize}
    \item The historical day-ahead prices of the previous three days and seven days ago: \(p_{d-1}, p_{d-2},p_{d-3},p_{d-7}\)
    \item The day-ahead forecasts of the exogenous variables considered from day \(d\), \(d-1\) and \(d-7\): \(x^i_d=[x^i_{d,1},...,x^i_{d,24}]\), \(x^i_{d-1}=[x^i_{d-1,1},...,x^i_{d-1,24}]\), \(x^i_{d-7}=[x^i_{d-7,1},...,x^i_{d-7,24}]\) where \(i\) is the exogenous variable in question. 
    \item A binary dummy variable that represents the day of the week: \(z=[z_{d,1},...,z_{d,7}]\) where each parameter is 0 except the one corresponding to the predicted day of the week. For example, in a Tuesday the variable would be \(z=[0,1,0,0,0,0,0]\)
\end{itemize}

The LEAR equation's coefficients will be analysed since they indicate the weight that the LEAR model assigns to each feature. Of course, the actual contribution will be the product between these coefficients and the feature value itself. This model was trained using a calibration window of one year and then used to predict hourly prices from the beginning of 2016 until the end of 2021. The input features are normalized before being used by the forecaster. The actual implementation borrows heavily from the epftoolbox, outsourced in \cite{b2}, but supplements the implemented price forecaster with more input features, as explained previously.

\section{Results and Discussion}

\begin{figure}[htbp]
\centerline{\includegraphics[width=0.45\textwidth]{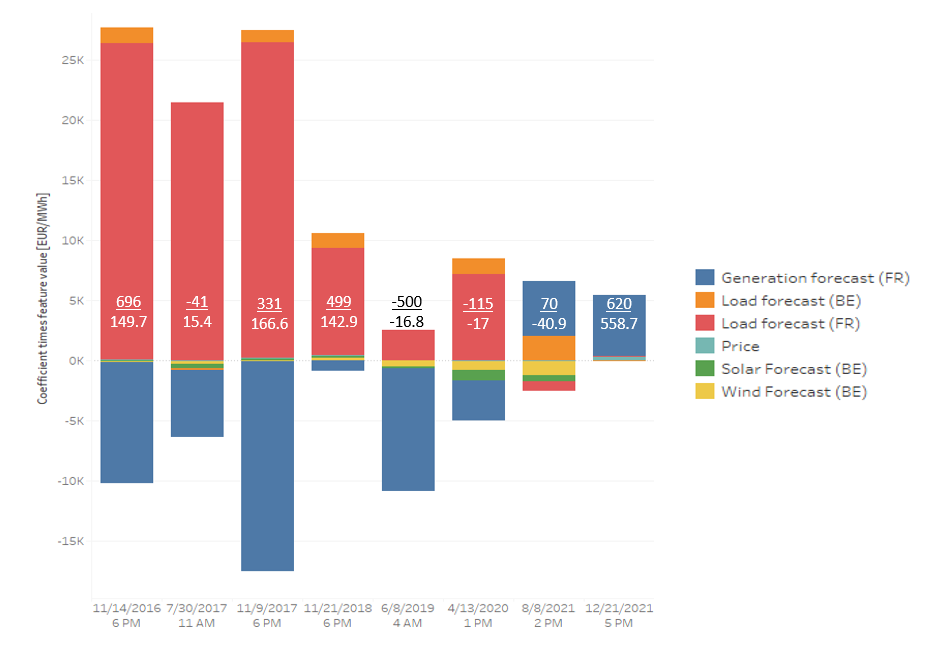}}
\caption{Product between LEAR model's coefficients and feature value, which the sum plus the b-value of the regression for each hour equals the predicted price.}
\label{Paper4}
\end{figure}

Before analyzing the physical consistency of the forecast model, it is important to understand that the LEAR model fails to predict accurate prices when there are: (1) negative or unusually low electricity prices and (2) unusually high prices. Outside of these two scenarios, the model is significantly accurate. This is illustrated in Fig.~\ref{Paper3}. In Fig.~\ref{Paper4} it is possible to observe the product between the model's coefficients and the feature value of four hours with spikes and four hours with negative prices. The real price (underlined) and the predicted price are also presented in each bar. These eight hours were chosen at random to illustrate the importance that the LEAR model attributes to each feature.

\subsection{Negative prices}
The model, in general, seems incapable of modelling negative electricity prices accurately. Negative electricity prices in Belgium are often due to a high energy generation surplus (e.g. through renewable sources) combined with low demand. This implies that the model should learn a negative causal link between (predicted) renewable energy generation and electricity prices. However, for most of the experimental period, the LEAR model places rather low importance to both solar and wind. In fact, the features with the most significant influence include the French load and generation forecasts. Likewise, the coefficients assigned to French electricity generation are supposed to be negative since larger imports would decrease the price of electricity in Belgium. However, this is not evidenced during 2021, when prices crashed in the post-Covid lockdown periods. 
Therefore, for the negative prices to be more accurately predicted, renewable energy generation should have a more considerable influence and the load forecast in France a smaller one.

\subsection{High prices}
For high prices, the model makes accurate predictions when the prices of the previous days are high as well, owing to the autoregressive component. However, when there is a spike in electricity prices, i.e., an isolated high price, the LEAR model fails to predict this spike, and instead predicts much lower price values. 
Regarding the spikes, the features that most influence the prediction of the LEAR model are the French generation and load forecasts. Since most of the analysed spikes take place during winter evenings, it makes sense for the Belgian wind generation forecast to be a key determinant. However, this would not necessarily fix under-predictions, since the predictions are often lower than the real price.
What would make the predictions more accurate is assigning greater importance in this case to the load forecast in Belgium and the load forecast in France for the last two days.

\section{Conclusion}

\begin{figure}[htbp]
\centerline{\includegraphics[width=0.45\textwidth]{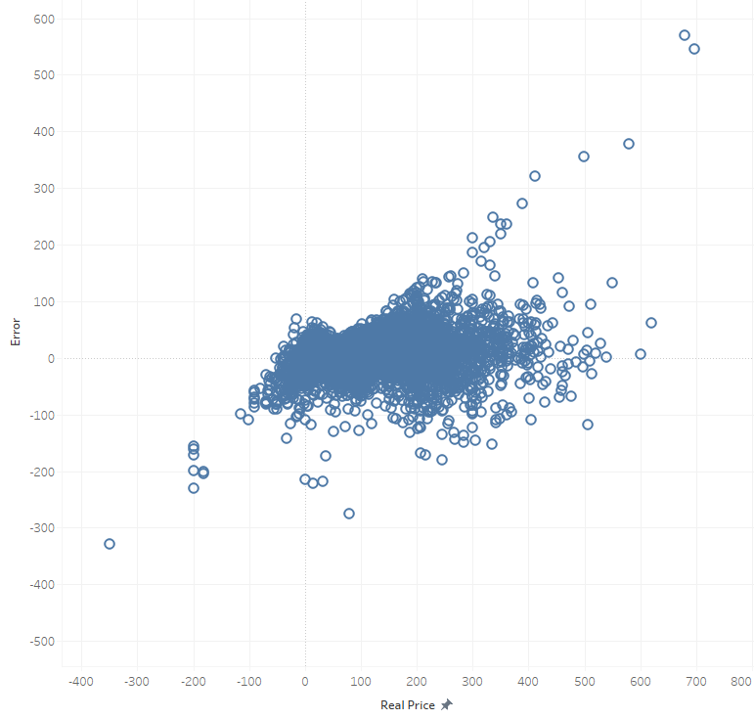}}
\caption{Scatter plot of the error (difference between the real and predicted price) vs real price.}
\label{Paper3}
\end{figure}

This paper shows the importance of analyzing a model's accuracy in greater detail, and studying the forecasting models' predictions for underlying causality. Thus, it is evident from this brief analysis that the LEAR model is quite accurate, but it fails during extreme conditions, as seen in Fig.~\ref{Paper3}. This should be foreseeable by a domain expert, but is not necessarily something the model generalizes well on. 
Moreover, with the pandemic, the model's accuracy appears to have decreased.

Creating new models is not always the solution to a more reliable forecast algorithm, but understanding where the model fails and consequently ensuring that the importance given to input features is physically consistent (and causally accurate). Therefore, more work should be put into understanding the models and improving their interpretability. The paper includes preliminary ideas for how the models could have performed more accurately, but these suggestions are limited by the issue of multicollinearity of features, which makes it important to interpret the model's behaviour. A more thorough analysis can address this, in addition to also investigating how other state of the art models, such as deep neural networks, perform on this metric.

\section*{Appendix}

\begin{thebibliography}{00}

\bibitem{b1} Catalão, J.P.S., Mariano, S.J.P.S., Mendes, V.M.F., and Ferreira, L.A.F.M. “Short-term electricity prices forecasting in a competitive market: A neural network approach”. In: Electric Power Systems Research 77.10 (2007),
pp. 1297–1304. ISSN: 0378-7796. DOI: https://doi.org/10.1016/j.epsr.
2006.09.022. URL: https://www.sciencedirect.com/science/article/pii/
S0378779606002422.

\bibitem{b2} Lago, J., Marcjasz, G., De Schutter, B., \& Weron, R. (2021). EPFTOOLBOX: The first open-access PYTHON library for driving research in electricity price forecasting (EPF). WORMS Software (WORking papers in Management Science Software).

\bibitem{b5} ENTSO-E. ENTSO-E Transparency Platform. 2022. URL: https://transparency.entsoe.eu/dashboard/show.


\bibitem{b6} Rte. View data published by RTE. 2022. URL: https://www.services- rte.com/en/view-data-published-by-rte.html.


\bibitem{b7} Uniejewski, Bartosz, Nowotarski, Jakub, and Weron, Rafał. “Automated Variable Selection and Shrinkage for Day-Ahead Electricity Price Forecasting”. In: Energies 9.8 (2016). ISSN: 1996-1073. DOI: 10 . 3390 / en9080621. URL: https://www.mdpi.com/1996-1073/9/8/621
\end{thebibliography}
\end{document}